\documentclass[letterpaper, 10 pt, conference]{ieeeconf}
\usepackage{graphicx} 
\usepackage{cite}
\usepackage{mathptmx}
\usepackage{times}
\usepackage{url}
\usepackage{amsfonts, amsmath}
\usepackage{bm}
\usepackage{amssymb}
\usepackage{interval}
\usepackage{tabularx}
\usepackage{multirow}
\usepackage{textcomp}
\usepackage{xcolor}
\usepackage{tikz}
\usepackage{pgf-pie}
\usepackage{siunitx}
\usepackage{float}

\IEEEoverridecommandlockouts
\title{\LARGE \bf
Autonomous Integration of Bench-Top Wet Lab Equipment
}

\author{Zachary Logan$^{1}$, Kam Undieh$^{2}$, and Mohammad Goli$^{3}$
\thanks{$^{1}$ Noblis Autonomous Systems Research Center Reston, Virginia {\tt\small Zach.Logan@noblis.org}}%
\thanks{$^{2}$ Noblis Autonomous Systems Research Center Reston, Virginia {\tt\small Mohammad.Goli@noblis.org}}%
\thanks{$^{3}$ Noblis Autonomous Systems Research Center Reston, Virginia {\tt\small Kam.Undieh@noblis.org}}%
}

\begin{document}

\maketitle
\thispagestyle{empty}
\pagestyle{empty}

\begin{abstract}
    Laboratory automation is an expensive and complicated endeavor with limited inflexible options for small-scale labs. We develop a prototype system for tending to a bench-top centrifuge using computer vision methods for color detection and circular Hough Transforms to detect and localize centrifuge buckets. Initial results show that the prototype is capable of automating the usage of regular bench-top lab equipment.
\end{abstract}

\section{Introduction}
Total laboratory automation (TLA) has existed for a number of decades and has improved the reproducibility~\cite{Hawker2000, Holland2020}, efficiency~\cite{Da2016, Hawker2000}, and safety~\cite{Movsisyan2016, Caragher2017} of the laboratory. TLA has also reduced the costs~\cite{Archetti2017} and number of man hours needed to complete a lab protocol~\cite{Holland2020}. Although, laboratory automation brings many benefits, there are still a number of obstacles limiting its use. The largest barriers to implementing laboratory automation are the cost and the lack of flexibility of the equipment. Automation solutions are expensive and are usually outside of the reach of smaller research labs. For example, Tecan's Freedom EVO, Hamilton's Star System, and Tap Biosystems's Compact Select can cost up to and over one million U.S. dollars~\cite{Storrs2013}. 

The majority of low-cost commercial automation systems that do exist are limited to only performing a singular task. There is also a lack of interoperability between laboratory devices~\cite{Rupp2022}, making it difficult to use a variety of equipment from different vendors in the same protocol. There are a few initiatives such as Standardization in Lab Automation,or SiLA,~\cite{Bar2012} and the Laboratory automation Plug \& Play, or LAPP, Framework~\cite{Wolf2022} that are working to address this lack of communication between lab devices; however, not all devices out on the market are SILA compliant and many manufactures still develop devices using proprietary systems. There are some low-cost options for performing laboratory automation such as Andrews+, Opentrons, and Hamilton; however, they lack the ability to integrate existing lab equipment such as incubators, centrifuges, and shakers, greatly limiting the flexibility of these types of options. 

We seek to create an autonomous system capable of interfacing with a variety of different inexpensive bench-top lab equipment to both simplify and facilitate the adoption and automation of new or early-stage lab procedures, reduce workloads for lab technicians, and remove people from working with and in hazardous environments and equipment. In this paper, we focus on automating the centrifugation process by using computer vision to detect the possible locations of test tubes in a conventional bench-top centrifuge and identify if a location contains a test tube or if it is an empty space that can accept a test tube.
\begin{itemize}
\item We construct a prototype robotic system that uses computer vision to insert and remove test tubes from a bench-top centrifuge without human intervention.
\item We experimentally verify the efficacy of the test tube detection and localization system by running it for 20 trials under various lightning conditions.
\item We experimentally verify the efficacy and robustness of the control system, the test tube detection, and the localization system by having it perform 40 randomized insertion and removal operations.
\end{itemize}
\begin{figure}[!ht]
    \includegraphics[width = \columnwidth]{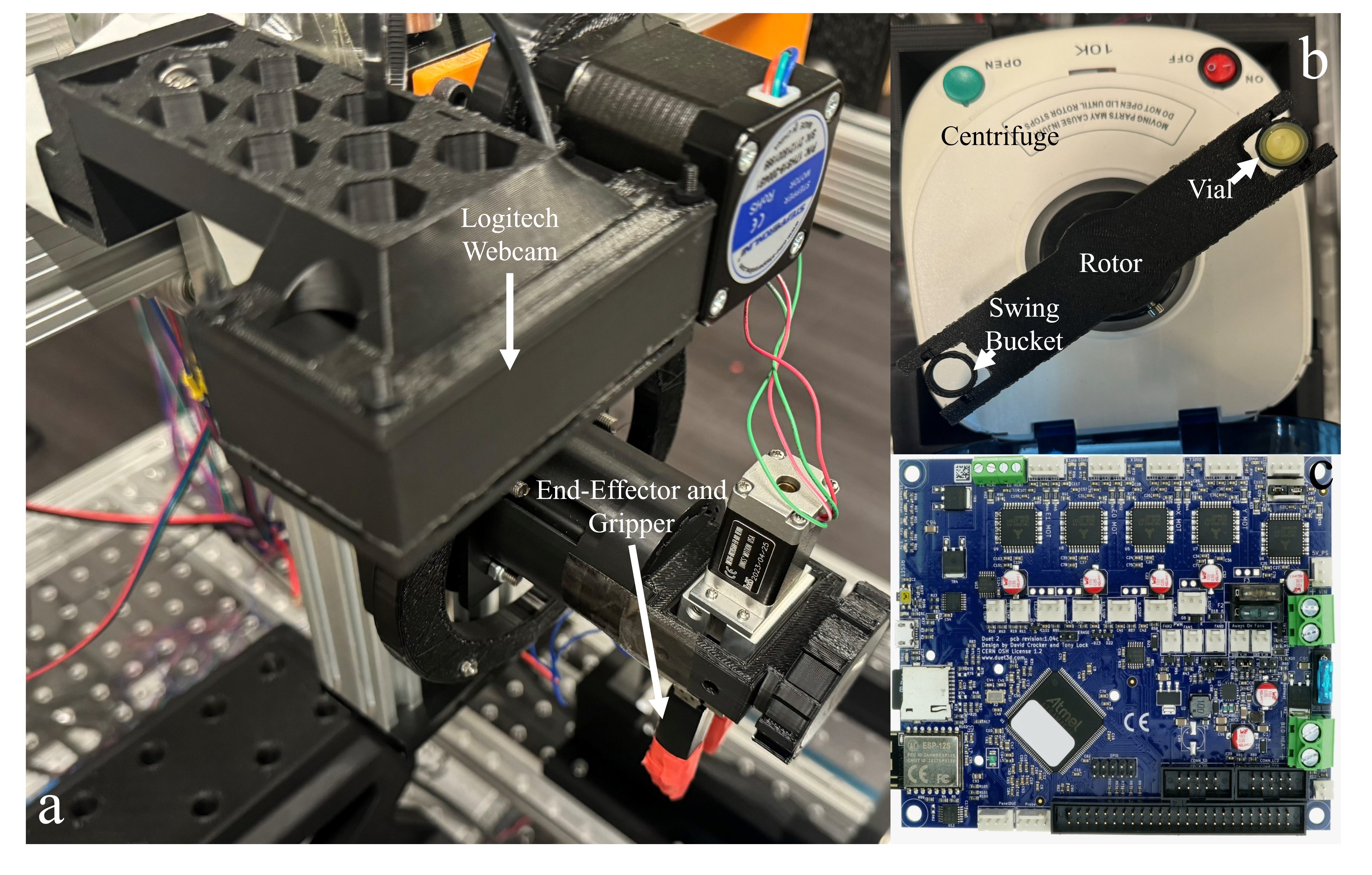}
    \caption{To automate a bench-top centrifuge, we used a 3D printer style gantry due to its simple control nature. We equipped the gantry system with a one degree of freedom end-effector and a Logitech C270 Webcam as shown in a). We chose to use the MiniPCR Bio Gyro Centrifuge as shown in b) because of its low-cost and wide availability. We equipped the centrifuge with a custom 3D printed swing bucket rotor to meet the required relative centrifugal force used in the prototypes DNA extraction protocol. To control the motors and sensors within the gantry system, we used the Duet 2 Ethernet micro-controller board~\cite{Duet2}, shown in c), because of its vast use in the 3D printer space and its expansive documentation.}
    \label{fig:experimental_setup}
    \vspace{-16pt}
\end{figure}

\section{Related Work}
Godolphin et al.~\cite{Godolphin1990} worked on a prototype to automate the centrifuging process by having a robot load and unload a conventional centrifuge. The prototype was created using a Micro-robot Type RM-501 with a dynamic centrifuge balancing algorithm. While the prototype was successfully able to automate the process, they encountered multiple issues with the solution. There was a two percent error rate caused by difficulty in gripping and manipulating the test tubes. They also found that the speed was insufficient for a hospital environment. They concluded that conventional centrifuges are not a good candidate for robot automation due to the batch-wise nature of insertion and the unpredictable position of test tube slots. Rupp et al.~\cite{Rupp2022} created a simple automation solution that can be used across a wide range of applications. The design used a Dobot Magician and the scripting software AutoIt to perform basic laboratory tasks like pipetting, autosampling, and inoculation. While this design performed well in the tested tasks, it could not integrate lab equipment where the position of the test tubes is not pre-defined.

To support dynamic test tube locations, computer vision can be used to perform a procedure known as object detection. Object detection is a number of extensively studied computer vision techniques used to classify and localize an object within an image. There are two primary approaches to object detection: the classical or traditional techniques and the deep learning-based computer vision techniques. Some of the classical or traditional computer vision techniques for performing object detection are corner detection~\cite{Moravec1980, Harris1988}, edge detection~\cite{Canny1986}, Hough Transforms~\cite{Hough1959, Ballard1981}, and feature matching algorithms such as Scale-invariant Feature Transform (SIFT)~\cite{Lowe1999}, Speeded-up Robust Features (SURF)~\cite{Bay2006}, and Oriented FAST and BRIEF (ORB)~\cite{Rublee2011}. Deep learning computer vision methods use trained neural networks to perform object detection. Some of the deep learning-based techniques are Fully Convolutional Networks (FCN)~\cite{Long2015}, U-net~\cite{Ronneberger2015}, R-CNN~\cite{Girshick2014}, Fast-RCNN~\cite{Girshick2015}, You Only Look Once~\cite{Redmon2016}, and Faster-RCNN~\cite{Ren2015}. 

Compared to classical methods, deep learning-based computer vision methods tend to provide results with greater accuracy, are easier to implement, and are more flexible~\cite{OMahony2020}. In traditional object detection the computer vision engineer has to manually determine what features best represent an object, but deep learning methods can be trained to learn features without explicit feature definitions. However, deep learning requires a large amount of data and computational power for training. While deep learning has been found to outperform traditional methods~\cite{Plaksyvyi2023}, there are still many cases where traditional methods are more effective. Traditional methods tend to be much more efficient when the number of object classes is very small or the objects have very clear differences, such as color. Although deep learning methods can still be used for these simple problems, a large amount of training data needs to be collected, and if the data is not collected and developed properly, it can cause the deep learning method to perform poorly~\cite{OMahony2020}. 

In our system, we determine possible test tube locations within the centrifuge and identify if those locations are empty or occupied. The locations for test tubes are all the same shape and size with very little variation, and there are only two possible object classes empty and present test tube. Because the problem is simple, we have chosen to perform object detection with classical computer vision methods. Due to the circular shape of both the test tube holders and the tubes themselves, we use the circular Hough Transform to perform the detection of the test tube holders. Since test tubes come in different types of colors, we chose to use a simple color threshold to determine the presences of test tubes in the system. 

\section{Methodology}
The prototype consists of two parts: the physical robot system and the computer vision software. The computer vision system localizes the buckets and test tubes and sends the positions to the robot system which uses those positions to generate the necessary movement commands to remove or insert test tubes into the centrifuge. We used the prototype to conduct three experiments that test the accuracy of the computer vision system, the localization accuracy of the full system, and the total run time of the different procedures.  

\subsection*{Prototype}

\subsubsection*{Robot System}
The robot system is a three-axis gantry with a one degree of freedom end-effector powered by NEMA 17 motors with 1.8 degrees per step. The three primary axes were setup to resemble a Cartesian 3D printer to simplify the control.

\paragraph*{Camera Type and Specs}
The experimental apparatus used a Logitech C270 Webcam mounted at an offset from the end-effector as shown in Figure~\ref{fig:experimental_setup}a. The webcam had a maximum resolution of 720p at 30 frames per second and was calibrated using OpenCV~\cite{opencv_library} in Python. 

\paragraph*{End-Effector}
The end-effector, as shown in Figure \ref{fig:experimental_setup}a, has one degree of freedom that allowed for rotation about the gantry systems x axis. The end-effector uses an array of different tools that are attached using an electromagnet. The tool shown in Figure \ref{fig:experimental_setup}a is a gripper tool containing two 3D printed fingers. The fingers are wrapped in tape to increase the friction between the test tubes and the fingers by making the surface pliable. 

\paragraph*{Control System}
The entire system was controlled using a Jetson Nano running Ubuntu 18.04 and a Duet 2 Ethernet board~\cite{Duet2} board running RepRap firmware~\cite{RepRap}, as shown in Figure~\ref{fig:experimental_setup}c. The Duet 2 Ethernet board is developed by Duet3D for controlling motors and other equipment in a 3D printer using G-code, which is a widely used for computer numerical control in 3D printing and machining equipment. The Jetson Nano controls the gantry by sending G-code commands to the Duet 2 using serial communication with the same methodology used by popular G-code sending software such as OctoPrint and Printrun by PronerFace. 

\subsubsection*{Bench-top Centrifuge}
In this experiment, we used a miniPCR Bio Gyro microcentrifuge~\cite{GyroCentrifuge}, shown in Figure~\ref{fig:experimental_setup}b. This bench-top centrifuge comes with two interchangeable rotors for holding a maximum of six test tubes in sizes of 0.2, 0.5, 1.5, or 2.0 milliliters or alternatively a maximum of sixteen 0.2 milliliter PCR tubes. The centrifuge operates at 100-240 volts AC with a fixed rotation speed of 10,000 revolutions per minute (RPM) and a maximum relative centrifugal force (RCF) of 4800 x g. The original rotor could not provide the RCF for the DNA extraction protocol utilized by this prototype. We designed a custom rotor with a larger radius to allow the system to reach a higher RCF while keeping the RPM at a constant value. To make access to the centrifuge by the gantry consistent, the custom rotor was designed to use swing-out style buckets as shown in Figure~\ref{fig:experimental_setup}b. The use of swing-out style buckets made it so that when the centrifuge is not in operation, the test tubes are mostly vertical. This vertical position simplifies the placement of test tubes. No modifications to the bench-top centrifuge, like direct control of the motor, are required to ensure that located test tubes and buckets are accessible to the robot.   

\subsection*{Computer Vision Setup}
 
\subsubsection*{Computer Vision Analysis}
\begin{figure}[!h]
    \centering
    \includegraphics[width=\columnwidth]{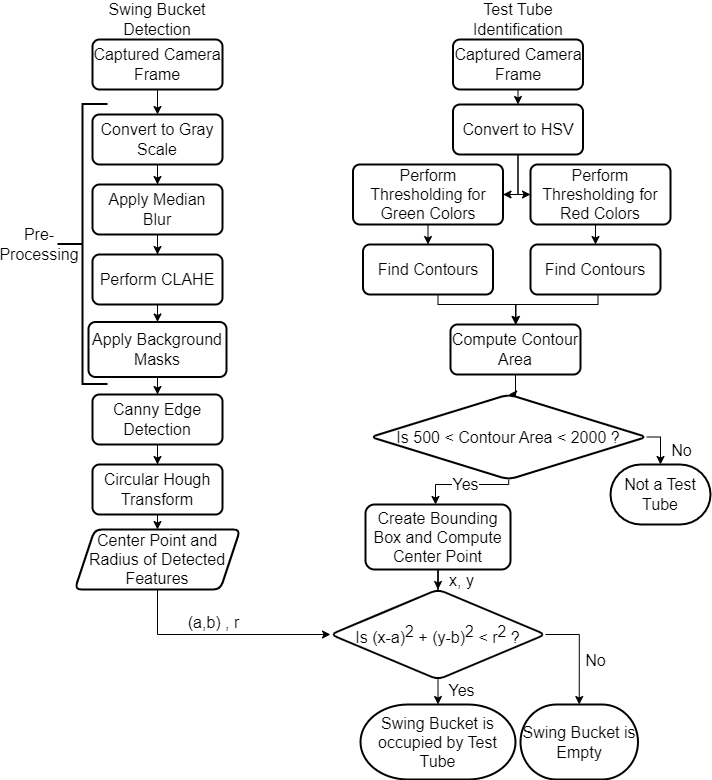}
    \caption{Diagram of the operations and decisions performed by the computer vision system.}
    \label{fig:computer_vision_diagram}
    \vspace{-10pt}
\end{figure}

Before the robot system can interact with the bench-top centrifuge, it first needs the location of each possible target and whether that target is empty or contains a test tube. In our system, there are only two possible targets on the custom centrifuge rotor. The computer vision system, outlined in Figure~\ref{fig:computer_vision_diagram}, is broken up into two steps: detect the swing buckets and identify if the detected feature is an open swing bucket or one containing a test tube. To begin the image analysis, a picture of the centrifuge rotor is taken using the webcam.

\paragraph{Swing Bucket Detection}
Before attempting feature detection, we first prepare the image with a few pre-processing steps. The original image is converted to gray-scale, and then a median blur with a 9x9 kernel is applied. We chose to use median blur to reduce the noise in the image while still retaining edge quality. A contrast limited adaptive histogram equalization (CLAHE) is applied to the image with clip value of 5 and kernel size of 8x8. The CLAHE increases the contrast between the dark rotor and its environment, to allow for a larger variation in lighting conditions. We then apply two masks to the CLAHE image to block out the center of the centrifuge rotor and the background area past the outer most edge of the rotor, where there are no relevant objects to be detected. Together, these masks reduce the chance that an extraneous object is detected.

After pre-processing is complete, the system detects the swing buckets by looking for the circular central hole. This detection was performed using OpenCV's Hough Circle Transform~\cite{opencv_library}. Hough Circle transform starts by performing a Canny Edge detection, then uses the edges to detect circular features using the Hough Gradient method~\cite{Yuen1990}. For the Canny Edge detection we set the lower hysteresis threshold at 20 and the upper hysteresis threshold at 120. Then for the Hough Circle Transform, we set the accumulator resolution ratio to be 1.5, the minimum distance between detected circles in the image frame to 200, the minimum circle radius in the image frame as 30, and the maximum circle radius in the image frame to 35. Through the Hough Transform, the center point and radius, in image coordinates, of each circular feature is found. 
 
\paragraph{Test Tube Identification}
After swing bucket detection, we identify if each swing bucket is empty or occupied. In our setup, we use color detection to identify if buckets are occupied with a test tube or empty. The captured camera image is converted into the HSV color space, and masks---the same masks used by the Swing Bucket Detection described earlier---are applied to block out part of the image beyond the centrifuge rotor. We generate three color masks to capture the colors of red, orange, green, yellow, pink, and purple, colors found in standard test tubes. We exclude shades of blue and gray as they were too similar in color to the centrifuge base. We then apply the color masks to the HSV image and increase the object area using the dilate function of OpenCV with a 5x5 kernel. We then use OpenCV's findContours~\cite{Suzuki1985, opencv_library} function to generate curves surrounding any objects found in the color masks. To help filter out invalid detections, we only accept contours with an area greater than 500 pixels and less than 2000 pixels. For each contour, we generate a bounding box that contains it. At this state each accepted contour is at a possible test tube location. To determine if the contour is a test tube, we test whether the center point of the bounding box lies within one of the circles detected by the Hough Circle Transform. If the center of the bounding box lies within any of the detected circles, then the corresponding swing bucket is said to be populated by a test tube. If none of the centers from the bounding boxes lie within a detected circle, then the corresponding swing bucket is said to be empty.

\begin{figure}[!h]
    \centering
    \includegraphics[width=\columnwidth]{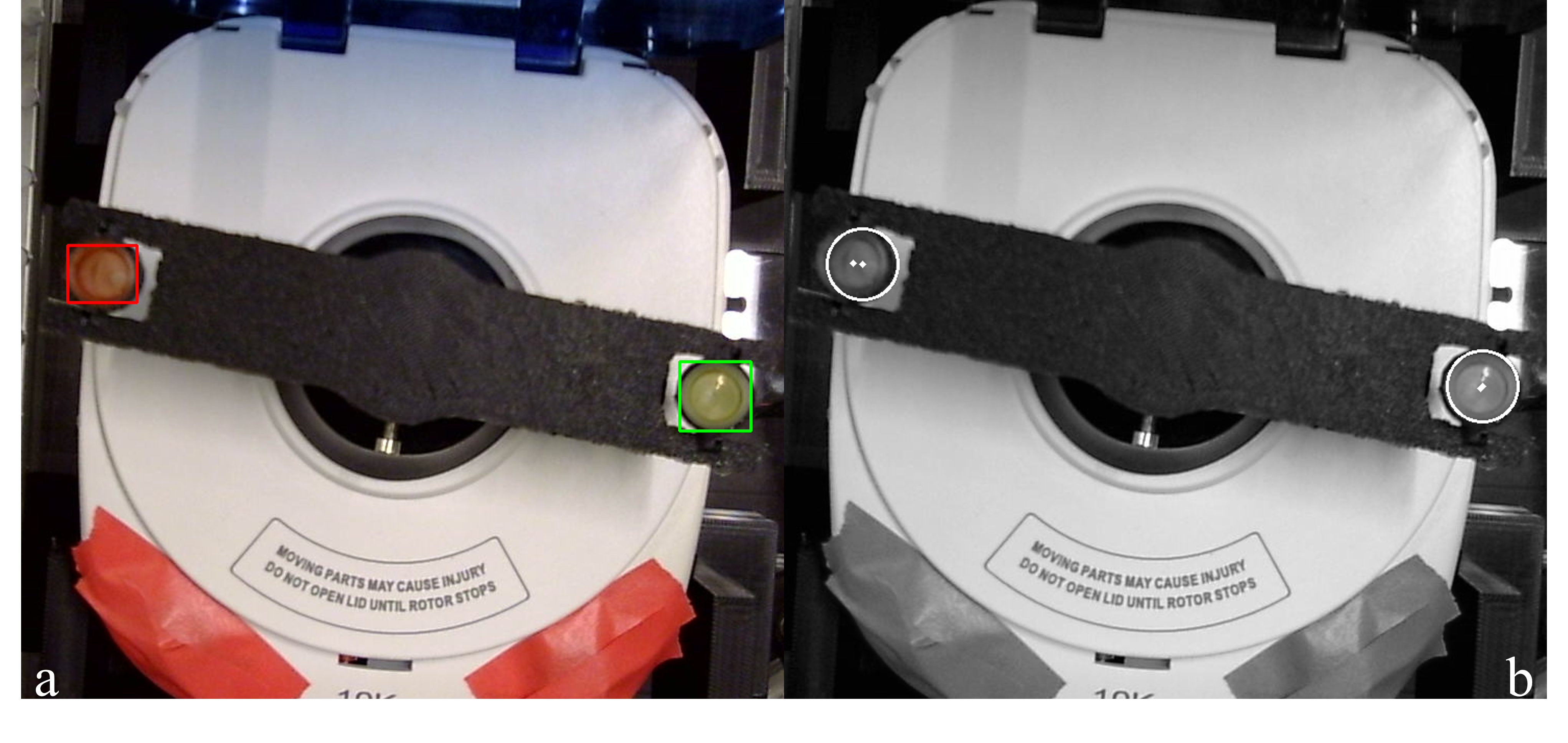}
    \caption{These images represent a sample output from the computer vision analysis system where the buckets are occupied by a yellow and orange test tube. a) This image shows the final bounding rectangles found after computing the contours found from the color detection system. b) This image shows the circular features and their center points found using the Hough Circle Transform, alongside the center points of the bounding rectangles found by the color detection.}
    \label{fig:computer_vision_output}
    \vspace{-18pt}
\end{figure}

A sample output of the computer vision analysis can be seen in Figure~\ref{fig:computer_vision_output}. Figure~\ref{fig:computer_vision_output}a shows the bounding rectangles outputted from the color-based contour detection. The output from the circular Hough Transform is shown in Figure~\ref{fig:computer_vision_output}b along with the center points of the bounding rectangles shown in Figure~\ref{fig:computer_vision_output}a.
\paragraph{Transform from Image Coordinates to World Coordinates}
At this point in the detection system, all positional information has been processed in the image frame of reference. In order to determine the true location of the swing buckets, we transform image pixel positions to real world positions. We perform the transform by computing the angle of the vector going from the known position of the center of the centrifuge rotor to the center of a circle found by the computer vision analysis in image coordinates. Then using this angle and the radius of the centrifuge rotor in millimeters, we can compute the X and Y position of the detected object in the centrifuge's coordinate frame. We defined the centrifuge coordinate frame to have an origin located at the center of the rotor. Lastly, we translate from the centrifuge's coordinate frame to the gantry's coordinate frame. Now that the position of the detected object is in the gantry's coordinate, it can be converted into the various G-code commands used by the Duet 2 control board to move the robot system to the indicated location. 

\subsection*{Experiments}
To determine if the prototype is capable of interacting with the bench-top centrifuge in place of an actual person, we devised experiments that test the system in each of the steps a human would need to take to interact with the centrifuge: detection and identification of the correct test tube or test tube holder to interact with, moving to the location of the holder or test tube, and inserting or removing a test tube from the holder. For all of the experiments, there are two buckets and up to a maximum of two test tubes. 

\subsubsection{Experiment 1: Detection and Identification Accuracy}
In this experiment, the arm attempts to identify holes and test tubes in the centrifuge. For twenty trials, we ran the system and recorded the number of times the system correctly identifies the right number of holes and test tubes in the centrifuge with varying environmental conditions. For each trial, the system starts with the camera hovering above the centrifuge and then a random number of test tubes are placed in random locations. After running the trials, we note how many features were detected, how many test tubes were identified, and how many empty buckets were identified. For each failure we note the cause. 

\subsubsection{Experiment 2: Localization Accuracy}
In this experiment, we ran forty trials split evenly between test tube insertion and removal. For each trial, the gantry starts by hovering above the centrifuge, then moves to center the camera above the centrifuge. Once the camera is above the centrifuge, the computer system captures an image of the rotor and analyzes it. Once the analysis is complete and the bucket and test tubes are localized into the gantry coordinate frame, the gantry attempts to perform an insertion or removal operation. During insertion, test tubes are handed to the gantry gripper by hand rather than picked up from another location. This allows insertion to be tested in isolation. We recorded the same information from the detection and identification experiment plus the success rate of performing the insertion and removal operations. On failure, we noted the observed cause. 

\subsubsection{Experiment 3: Run Time}
To determine if the system is fast enough to replace a manual setup, we recorded the time it takes to perform each operation. We measured the detection and identification process twenty times, test tube removal process twenty times, and test tube insertion process twenty times. The run time was determined using the execution time of the code including all pauses and motor actuation. 

\section{Results}
\subsection{Detection and Identification}
The computer vision system successfully detected and identified the buckets and test tubes in the centrifuge with only a five percent detection error. As shown by the data summarized in Table~\ref{tab:detect_vial}, the five percent detection error was caused by the computer vision system detecting a bucket that was not present. In addition, the computer vision system always---for every trial run---identified the correct number of test tubes.

\begin{table}[!ht]
    \centering
    \begin{tabularx}{\columnwidth}{c*{7}{>{\centering\arraybackslash}X}}
    & \multicolumn{6}{c}{Number of Test Tubes} \\
    & \multicolumn{2}{c}{0} & \multicolumn{2}{c}{1} & \multicolumn{2}{c}{2} \\
    \cline{2-7}
    Metric & Actual & Desired & Actual & Desired & Actual & Desired \\
    \hline
    Detection & 16 & 16 & 17 & 16 & 8 & 8 \\
    Identification & 0 & 0 & 8 & 8 & 8 & 8 \\
    \end{tabularx}
    \caption{Test Tube Detection and Identification Trial Data}
    \vspace{-24pt}
    \label{tab:detect_vial}
\end{table}
\subsection{Localization}

\subsubsection{Removal}
\begin{figure}[!h]
    \centering
    \begin{tikzpicture}[font=\footnotesize, scale=0.85]
      \pie[square,color={green!30, red!20, red!60, blue!20, blue!50}, text=legend]{60.0/Success, 15.0/Localization Failure, 5.0/Identification Failure, 5.0/Detection Failure, 15.0/Hardware Interference}
    \end{tikzpicture}
    \caption{The robot system successfully removed the test tube sixty percent of the time. Fifteen percent of the trials resulted in an error due to improper localization, where the gripper was not properly lined up with the test tube. A total of ten percent of trials resulted in an error due to a failure by the computer vision system. Half of the computer vision errors were caused by the Hough Transform detecting a non-existent swing bucket and the other half were caused by improper test tube occupancy identification. The last fifteen percent of trials failed due to hardware-related interference.}
    \label{fig:vial_removal}
    \vspace{-12pt}
\end{figure}
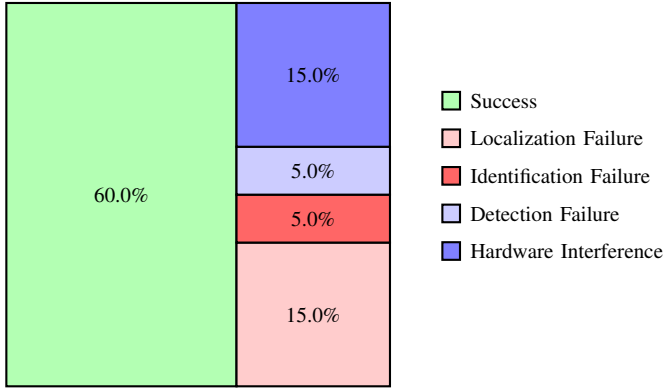

The system successfully removed vials from the centrifuge sixty percent of the time. Unsurprisingly, the primary causes of error, as shown in Figure~\ref{fig:vial_removal}, were improper localization and hardware related interference. Another result was that five percent of trials failed due to a detection error. Similar to the results from Experiment 1, the error was caused by the computer vision system detecting an extra swing bucket. This data is summarized in Table~\ref{tab:remove_vial}. A slight surprise was that five percent of trials failed due to improper test tube occupancy identification. 

\begin{table}[!h]
    \centering
    \begin{tabularx}{\columnwidth}{c*{7}{>{\centering\arraybackslash}X}}
    &\multicolumn{6}{c}{Number of Test Tubes} \\
    &\multicolumn{2}{c}{0} &\multicolumn{2}{c}{1} &\multicolumn{2}{c}{2} \\
    \cline{2-7}
    Metric & Actual & Desired & Actual & Desired & Actual & Desired \\
    \hline
    Detection & 10 & 10 & 17 & 16 & 14 & 14 \\
    Identification & 0 & 0 & 7 & 8 & 14 & 14\\
    Localization & 0 & 0 & 7 & 8 & 8 & 14 \\
    \end{tabularx}
    \caption{Test Tube Removal Trial Data}
    \label{tab:remove_vial}
    \vspace{-12pt}
\end{table}

\subsubsection{Insert}
\begin{figure}[!h]
    \centering
    \begin{tikzpicture}[font=\footnotesize, scale=0.85]
      \pie[square,color={green!30, red!20, red!60},text=legend]{75.0/Success, 5.0/Localization Failure, 20.0/Identification Failure}
    \end {tikzpicture}
    \caption{The robot system successfully inserted the test tubes seventy-five percent of the time. Five percent of the trials resulted in an error due to improper localization. Twenty percent of trials resulted in an error due to improper test tube occupancy identification.}
    \label{fig:vial_insert}
    \vspace{-12pt}
\end{figure}
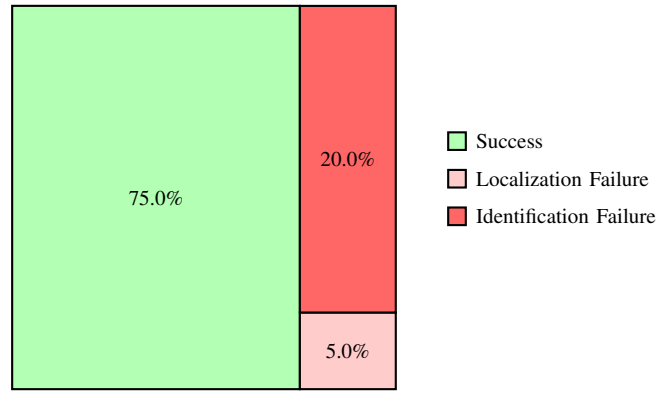

The system was more effective at inserting the test tubes into the centrifuge with a success rate of seventy-five percent as shown in Figure~\ref{fig:vial_insert}. To our surprise, the primary cause for an unsuccessful trial, twenty percent of failures, was due to improper identification of the test tubes. In contrast, localization errors were responsible for five percent of trial failures. As shown in the experimental data summarized in Table~\ref{tab:insert_vial}, the computer vision system identified fewer test tubes than were actually present in the trials, meaning that it was classifying swing buckets that were occupied by test tubes as empty. This change in failure type was surprising as there were no changes to the experimental setup when moving from the test tube removal experiments to the test tube insertion experiments. As shown in both Table~\ref{tab:insert_vial} and Figure~\ref{fig:vial_insert}, there were no failures caused by improper detection of the number of swing buckets. 

\begin{table}[!h]
    \centering
    \begin{tabularx}{\columnwidth}{c*{7}{>{\centering\arraybackslash}X}}
    &\multicolumn{6}{c}{Number of Test Tubes} \\
    &\multicolumn{2}{c}{0} &\multicolumn{2}{c}{1} &\multicolumn{2}{c}{2} \\
    \cline{2-7}
    Metric & Actual & Desired & Actual & Desired & Actual & Desired \\
    \hline
    Detection & 20 & 20 & 8 & 8 & 10 & 10 \\
    Identification & 0 & 0 & 2 & 4 & 8 & 10\\
    Localization & 21 & 22 & 4 & 4 & 0 & 0\\
    \end{tabularx}
    \caption{Test Tube Insertion Trial Data}
    \label{tab:insert_vial}
    \vspace{-16pt}
\end{table}

During the trials, we noted that the translucent test tubes on the right side of the image appeared far brighter than ones on the left side of the image. Further inspection yielded that there was a large amount of light glare present on these test tubes, causing their appearance in the camera image to be over-saturated. This over saturation reduced the amount of color information in camera images, causing the computer system to improperly identify the occupancy of a swing bucket. 

\subsubsection{Localization Discussion}
The difference in the number of failures caused by improper localization in the insertion and removal trials, was caused by the respective positional tolerances of the operations. To properly grasp and raise the test tube out of the bucket, the gripper required precise alignment. In contrast, inserting a test tube into the centrifuge required much less precision. The insertion operation only requires that the tip of the test tube is within the bucket during the insertion process. The difference in test tube identification provides us with a clear indication that there is both a minimum and maximum amount of ambient light that can be present. 

\subsection{Run Time}
As shown in Table~\ref{tab:run_time}, the computer vision system took approximately 10.2 seconds to complete. Surprisingly, the removal operation took longer to complete than insertion. On average the removal operation took 172 seconds to complete, while the insertion operation took 129 seconds to complete. The majority of the run time in both the removal and insertion came from the control codes serial communication protocol with the Duet 2 Ethernet board. Even with this bottleneck, the system may still be useful in its current state. The entire operation is automated and does not require human supervision, so even though the process is slow, the system frees up time that can be spent completing other tasks and assignments.

\begin{table}[!h]
    \centering
    \begin{tabularx}{\columnwidth}{c*{4}{>{\centering\arraybackslash}X}}
        Process & Present Test Tubes & Run Time (seconds) & Average Time (seconds) \\
        \hline
        \multirow{3}{4em}{Detection} & 0 & $8.260$ & \multirow{3}{4em}{$10.135$} \\
        & 1 & $11.301$ &\\
        & 2 & $11.551$ &\\
        \hline
        \multirow{3}{4em}{Insert} & 0 & $179.293$ & \multirow{3}{4em}{$129.166$}\\
        & 1 & $115.180$ &\\
        & 2 & $30.628$ &\\
        \hline
        \multirow{3}{4em}{Removal} & 0 & $22.016$ & \multirow{3}{4em}{$172.098$} \\
        & 1 & $141.778$ &\\
        & 2 & $311.582$ &\\
    \end{tabularx}
    \label{tab:run_time}
    \caption{Run Times for Test Tube Detection and Identification, Test Tube Removal and Test Tube Insertion}
    \vspace{-24pt}
\end{table}

\section{Conclusion}
In this paper we designed, developed, and tested a prototype system to automate a bench-top centrifuge, which has applications in reducing the cost and number of man hours for lab sample processing. We used the Hough Transform and color detection computer vision techniques to create an object detection system to determine the location of the centrifuge buckets and to identify if a bucket is empty or occupied by a test tube. In our results the automation system was able to detect and identify the buckets and test tubes with a high level of accuracy and successfully inserted test tubes into the centrifuge in seventy-five percent of trials.

Future work may focus on two limitations: the effect of ambient environmental light and the speed and precision of the Duet-based control. While the prototype was tested with varying amounts of ambient light, the testing was still performed in an indoor controlled environment. Further testing under a wider array of applications, such as outdoors or in locations with much less control of the ambient conditions, would be needed to allow for effective functionality. The speed of the current prototype, while sufficient for small labs with low throughput, is not at a level capable of handling high amounts of throughput in a standard workday. Further optimization for the communication between the main Jetson computer and the Duet micro-controller is needed in order to achieve faster control. The current control and physical setup also resulted in precision issues. For example, two sequential movement commands to identical coordinates caused the arm to move slightly. Further refinement of the control and operational hardware will allow for more precise position control and reduce localization errors.

\bibliography{lit_material}
\bibliographystyle{ieeetr}

\end{document}